\documentclass[letterpaper, 10 pt, conference]{ieeeconf}  

\IEEEoverridecommandlockouts                              

\usepackage{graphicx}
\usepackage{amsmath}
\usepackage{rotating}
\usepackage{subcaption}
\usepackage{threeparttable}
\usepackage[utf8]{inputenc}
\usepackage[hidelinks]{hyperref}
\usepackage{multirow}
\usepackage[english]{babel}
\usepackage{nomencl}
\makenomenclature 
\usepackage{booktabs}
\usepackage{tabu}
\usepackage{caption}
\usepackage{cite}
\usepackage{amssymb}
\usepackage{comment}
\usepackage{longtable}
\usepackage{footnote}
\usepackage[table,xcdraw]{xcolor}
\DeclareCaptionLabelFormat{andtable}{#1~#2  \& \tablename~\thetable}
\usepackage[linesnumbered,ruled,vlined]{algorithm2e}
\SetKwInput{KwInput}{Input}                
\SetKwInput{KwOutput}{Output}              
\usepackage{ltablex,array}

\hypersetup
{
 unicode=false,     
 pdftoolbar=true,    
 pdfmenubar=true,    
 pdffitwindow=false,   
 pdfstartview={FitH},  
 pdftitle={My title},  
 pdfauthor={Author},   
 pdfsubject={Subject},  
 pdfcreator={Creator},  
 pdfproducer={Producer}, 
 pdfkeywords={keywords}, 
 pdfnewwindow=true,   
 colorlinks=true,    
 linkcolor=red,     
 citecolor=blue,    
 filecolor=magenta,   
 urlcolor=cyan      
}

\title{\LARGE \bf
Investigating Personalized Driving Behaviors in Dilemma Zones: Analysis and Prediction of Stop-or-Go Decisions
}

\author{Ziye~Qin,~\IEEEmembership{Student Member,~IEEE,}
        Siyan~Li,~\IEEEmembership{Student Member,~IEEE,}
        Guoyuan~Wu,~\IEEEmembership{Senior~Member,~IEEE,}\\
        Matthew~J.~Barth,~\IEEEmembership{Fellow,~IEEE,}
        Amr~Abdelraouf,~\IEEEmembership{Member,~IEEE,}\\
        Rohit~Gupta,
        and~Kyungtae~Han,~\IEEEmembership{Senior~Member,~IEEE}
\thanks{Ziye Qin, Siyan Li, Guoyuan Wu, and  Matthew J. Barth are with the Bourns College of Engineering, Center for Environmental Research and Technology, University of California at Riverside, Riverside, CA 92507 USA.}
\thanks{Ziye Qin is also with the School of Transportation and Logistics, Southwest Jiaotong University, Chengdu, Sichuan, China.}
\thanks{Amr Abdelraouf, Rohit Gupta and Kyungtae Han are with the Toyota Motor North America R$\&$D, InfoTech Labs, Mountain View, CA
94043.}
}

\begin{document}
\maketitle
\thispagestyle{empty}
\pagestyle{empty}

\begin{abstract}
Dilemma zones at signalized intersections present a commonly occurring but unsolved challenge for both drivers and traffic operators. Onsets of the yellow lights prompt varied responses from different drivers: some may brake abruptly, compromising the ride comfort, while others may accelerate, increasing the risk of red-light violations and potential safety hazards. Such diversity in drivers’ stop-or-go decisions may result from not only surrounding traffic conditions, but also personalized driving behaviors. To this end, identifying personalized driving behaviors and integrating them into advanced driver assistance systems (ADAS) to mitigate the dilemma zone problem presents an intriguing scientific question. In this study, we employ a game engine-based (i.e., CARLA-enabled) driving simulator to collect high-resolution vehicle trajectories, incoming traffic signal phase and timing information, and stop-or-go decisions from four subject drivers in various scenarios. This approach allows us to analyze personalized driving behaviors in dilemma zones and develop a Personalized Transformer Encoder to predict individual drivers’ stop-or-go decisions. The results show that the Personalized Transformer Encoder improves the accuracy of predicting driver decision-making in the dilemma zone by 3.7\% to 12.6\% compared to the Generic Transformer Encoder, and by 16.8\% to 21.6\% over the binary logistic regression model.
\end{abstract}


\section{Introduction}
\label{sec:Intro}
\subsection{Motivation and Related Work}
 The concept of the dilemma zone, which encapsulates the stop-or-go dilemma drivers face when encountering a yellow light, was first introduced in 1960 \cite{gazis1960problem}. Over the decades, dilemma zones at signalized intersections continue to pose a significant challenge. A considerable number of intersection accidents have occurred due to red-light violations, resulting from drivers failing to cope with dilemma zones \cite{wei2024dilemma,chauhan2022analysing}. The advent of advanced intersection management systems (AIMS) has brought technologies such as the internet of things (IoT), adaptive traffic signal control, and advanced driver assistance systems (ADAS) into focus as a potential solution to mitigating the dilemma zone issue \cite{qin2024game, chauhan2022analysing,das2022traffic}. However, the dilemma zone remains an existential and thorny problem, exacerbated by the intricate traffic dynamics at intersections, and the heterogeneity of 
driving behaviors. For instance, various drivers may exhibit diametrically opposed responses, choosing either to ``stop" or ``go", when faced with identical dilemma zone scenarios.

Dilemma zones are typically classified into two main categories: Type I dilemma zones, characterized by situations where drivers find it neither feasible to comfortably stop at the stop-line nor safely pass through an intersection during the yellow-light. 
Type II dilemma zones, where drivers face indecision regarding whether to pass through or stop at the intersection. Many efforts have been made to mitigate the dilemma zone problem, broadly falling into two categories: initiatives concentrating on enhancements of infrastructure-based perception/communication/control and strategies aimed at integrating in-vehicle assistance technologies \cite{wei2024dilemma}. From the infrastructure perspective, strategies such as dynamic signal timing optimization \cite{gao2021coordinated}, green extension systems (GES) \cite{zegeer1978green} and advanced warning flasher (AWF) \cite{sayed1999advance} have been proven effective in reducing red-light running. Loop detectors and roadside sensors (e.g., camera, radar, and lidar) are deployed to collect vehicle information at intersections to help mitigate dilemma zones \cite{zimmerman2007additional,elmitiny2010classification}. Furthermore, in-vehicle assistance can provide a warning for upcoming yellow-light or red-light using human machine interface (HMI) \cite{wei2024dilemma}. With the increasing diversity of collected information, it is evident that factors influencing stop-or-go decision-making in the dilemma zone extend beyond external conditions (e.g., weather, road condition, and speed limit) to include inter-driver differences \cite{rakha2008modeling, wei2024dilemma}. Studies have identified wider dilemma zones for older drivers and a greater propensity for male drivers to choose ``go" in dilemma zones \cite{rakha2008modeling}. Driver experience and concentration levels are also significant factors \cite{ghanipoor2016modeling, papaioannou2007driver}. The multitude of factors affecting driving behaviors in the dilemma zone complicates the task of modeling such behavior accurately. Modeling personalized driving behaviors emerges as a promising approach to capturing the nuanced characteristics of individual drivers, driven by the increasing data availability and the potential to provide tailored driving assistance \cite{liao2023driver}. 

Studies employing inverse reinforcement learning (IRL) have demonstrated that personalized adaptive cruise control (P-ACC) significantly outperforms the intelligent driver model (IDM) in imitating the drivers' longitudinal behaviors \cite{zhao2022personalized}. In scenarios such as lane-changing and ramp-merging, deep learning models that incorporate personalized characteristics, such as CVAE-LSTM and DenseTNT, have shown promising results in predicting lane-changing decisions and vehicle trajectories  \cite{bao2021prediction,Li2024personalized}. Efforts to develop personalized lane-changing trajectory planning strategies, including the integration of support vector machine (SVM) based classifier with model predictive control (MPC) \cite{vallon2017machine}, and improved rapidly-exploring random tree (RRT) and B-spline \cite{huang2021personalized}, have also been made. Personalized ADAS has been designed to optimize travel time and fuel consumption at signalized intersections \cite{butakov2016personalized}. Considering the interaction with surrounding vehicles, temporal graph neural networks (GNNs) and long short-term memory (LSTM) are used for personalized trajectory prediction \cite{abdelraouf2023interactionaware}. However, the specific area of personalized driver behavior in the dilemma zone, which is crucially related to intersection safety, has not yet been adequately addressed.

In this paper, we aim to conduct an in-depth investigation of personalized driving behaviors in dilemma zones. To achieve this, we utilized a game engine-based (i.e., CARLA-enabled) driving simulator to collect the high-resolution vehicle trajectories, stop-or-go decisions, and incoming traffic signal phase and timing information as drivers encounter dilemma zones. We then analyzed these behavioral characteristics and proposed a deep learning model to predict critical stop-or-go decisions. 

This paper makes several contributions to the field: 
\begin{itemize}
    \item A effective CARLA-enabled driving simulator has been developed to collect drivers' stop-or-go decisions within the dilemma zone in real time. 
    \item This study conducts a analysis of heterogeneous driving behaviors, both among different drivers (inter-driver heterogeneity) and within trips by each individual driver (intra-driver heterogeneity), with the aim of facilitating the prediction of personalized driving behaviors in the dilemma zone. 
    \item To more precisely capture the driver's decision-making process, we refine the concept of time to stop-line by recalculating its value based on ``stop" or ``go" decisions, respectively.
    \item A Personalized Transformer Encoder is proposed to predict stop-or-go decisions. This model incorporates the driver's personal characteristics, enhancing predition accuracy by 3.7\% to 12.6\% compared to the Generic Transformer Encoder, and by 16.8\% to 21.6\% over the binary logistic regression model.
\end{itemize}

\subsection{Organization of the Paper}
The organization of this work is as follows: Section \ref{sec: pf} outlines the methodology for investigating personalized driving behaviors in dilemma zones. Section \ref{sec: data} presents a overall analysis of personalized driving data. Section \ref{sec: prediction} introduces a model for predicting personalized stop-or-go decisions. Finally, the conclusion and directions for future work are presented in Section \ref{sec: con}.

\section{Methodology}
\label{sec: pf}
\subsection{Work Flow of Investigating Personalized Driving Behaviors}
\begin{figure*}[ht]
\centering
	\includegraphics[width=0.9\linewidth]{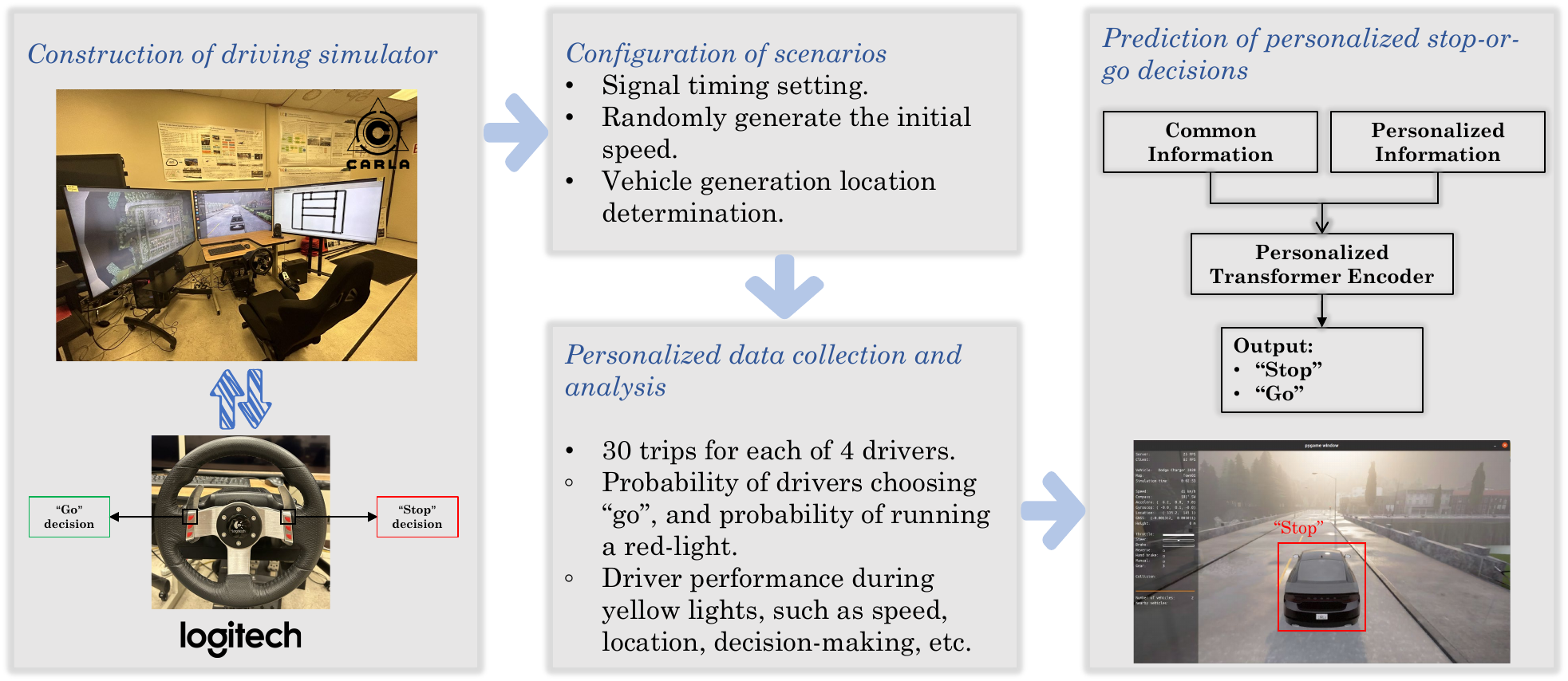}
\caption{Work flow of personalized driving behavior analysis and prediction.}
\label{Fig:sa}
\end{figure*}

In this section, we present a comprehensive work flow designed to investigate personalized driving behaviors within the dilemma zone, as depicted in Figure \ref{Fig:sa}. This architecture encompasses four key steps: the construction of CARLA-enabled driving simulator, the configuration of the simulation scenario, the collection and analysis of personalized driving behavior data, and the prediction of personalized stop-or-go decisions. Through the integration of these steps, the work flow is designed to simulate realistic driving environments, collect and analyze high-fidelity data on driver behavior, and employ this data to refine predictive models of drivers' decision. Each of these steps is discussed in more depth in the subsequent sections.

\subsection{Construction of CARLA-enabled Driving Simulator}
CARLA, an autonomous driving simulation platform, has become increasingly favored within the research community for its ability to create highly realistic traffic scenarios \cite{dosovitskiy2017carla}. Its appeal is further enhanced by its open-source nature and an accessible application programming interface (API), as documented in existing literature \cite{dosovitskiy2017carla,qin2024game}. Leveraging the capabilities of CARLA, we integrated the Logitech steering wheel kit as the input device for driver operations. To explicitly record drivers' decisions within the dilemma zone in real time, we assigned two buttons on the steering wheel to represent the ``stop" or ``go" choices, respectively.

\subsection{Configuration of the Simulation Scenario}
\begin{figure}[ht]
\centering
	\includegraphics[width=0.92\linewidth]{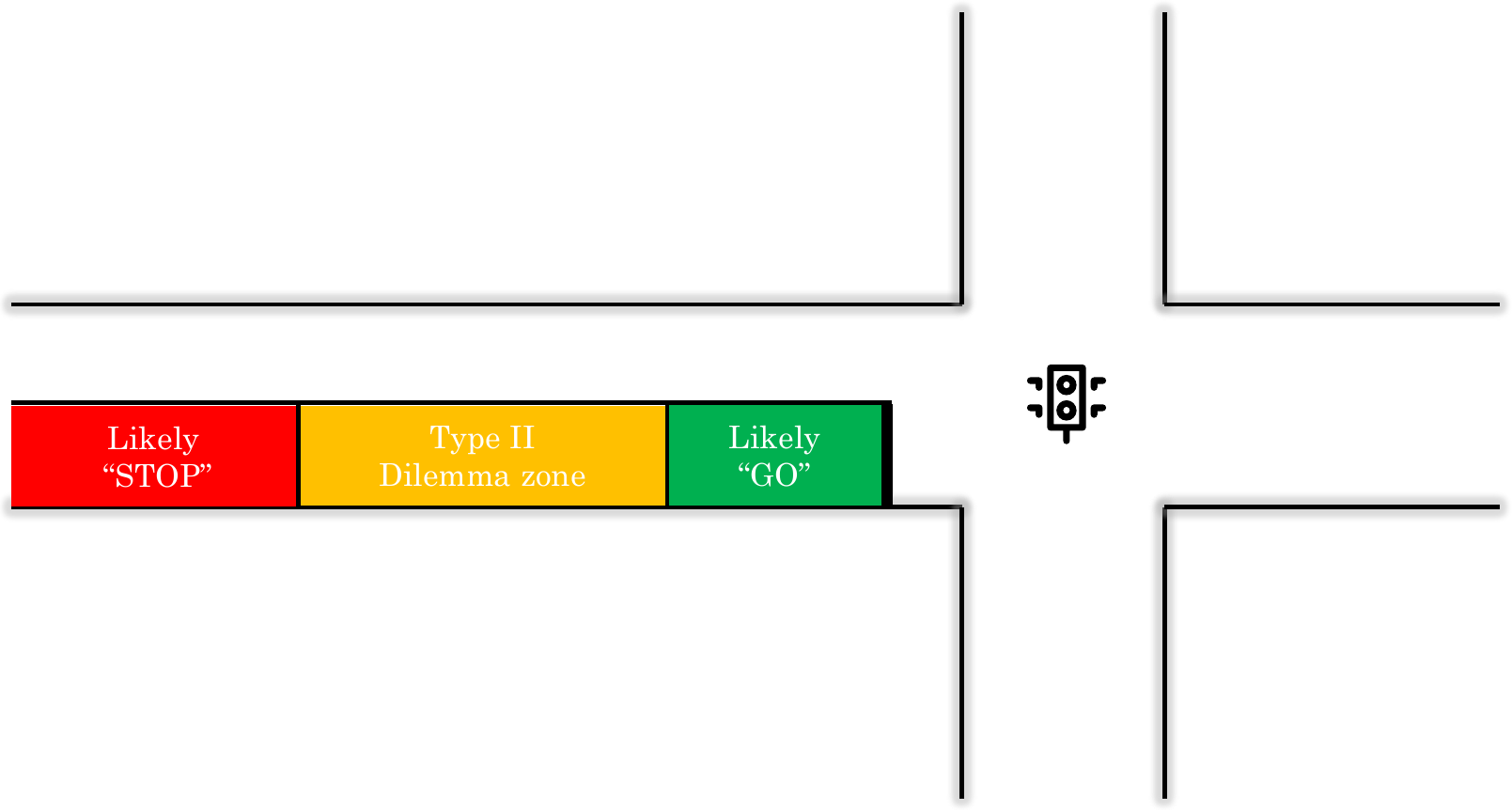}
\caption{Illustration of Type II dilemma zone.}
\label{Fig:dz}
\end{figure}

With respect to the scenario setup, to ensure a high likelihood of encountering a dilemma zone during each data collection session, we adopted a conventional definition for the range of dilemma zones. Specifically, the travel time from the start and end of Type II dilemma zone (as shown in Figure \ref{Fig:dz}) to the stop-line is about 5.5s and 2.5s, respectively \cite{bonneson2002intelligent, savolainen2016driver,wei2024dilemma}. At the beginning of each experiment, the initial vehicle speed is randomly generated to simulate different arrival patterns, while the vehicle position is determined in accordance with the above definition of dilemma zone. Additionally, the duration of the remaining green-light phase is appropriately adjusted to align with this condition.

\subsection{Collection of Personalized Driving Data}
Four volunteers were recruited as subject drivers, each executed 30 data collection sessions within designed dilemma zone scenarios, featuring randomized initial velocities and positions. Unlike conventional traffic research which always generalizes across many drivers, the study of personalized driving behavior focuses more on the variability within a specific driver, making it reasonable to assume that four volunteers would be appropriate here \cite{liao2023driver}. Such randomization is crucial for capturing a full spectrum of driving behaviors for an individual driver under different scenarios within the dilemma zone, thereby enhancing the robustness and applicability of our findings. To capture fine-grained driving behavior, the data collection frequency was set 50fps. This high-frequency data collection allows for the precise measurement of delicate changes in driving behavior, including subtle adjustments in speed and the timing of stop-or-go decisions. 

\subsection{Prediction of Personalized Stop-or-Go Decisions}
Based on our above simulation configuration and data analysis, we then proposed a Personalized Transformer Encoder that combined common information with personalized driver data to predict decisions in dilemma zones. The output contained two states: ``stop" and ``go", providing a comprehensive prediction results. This model uniquely considered each driver's historical behavior alongside current traffic signals and vehicle states, aiming to accurately forecast whether a driver will stop or proceed through an intersection. Followed by the results, we conducted experiments to compare the proposed model with the generic one, highlighting the significant advantage to use personalized model for dilemma zone decision-making process. This model supports the development of personalized ADAS, aiming to decrease the incidence of drivers facing dilemma zones to lessen the likelihood of drivers encountering dilemma zones, thereby improving safety and efficiency at intersections.

\section{Analysis Results of personalized driving data}
\label{sec: data}
\subsection{Overall Evaluation of Driver Behavior}
\begin{figure}[ht]
\centering
\begin{subfigure}{0.49\linewidth}
		\centering
		\includegraphics[width=1.0\linewidth]{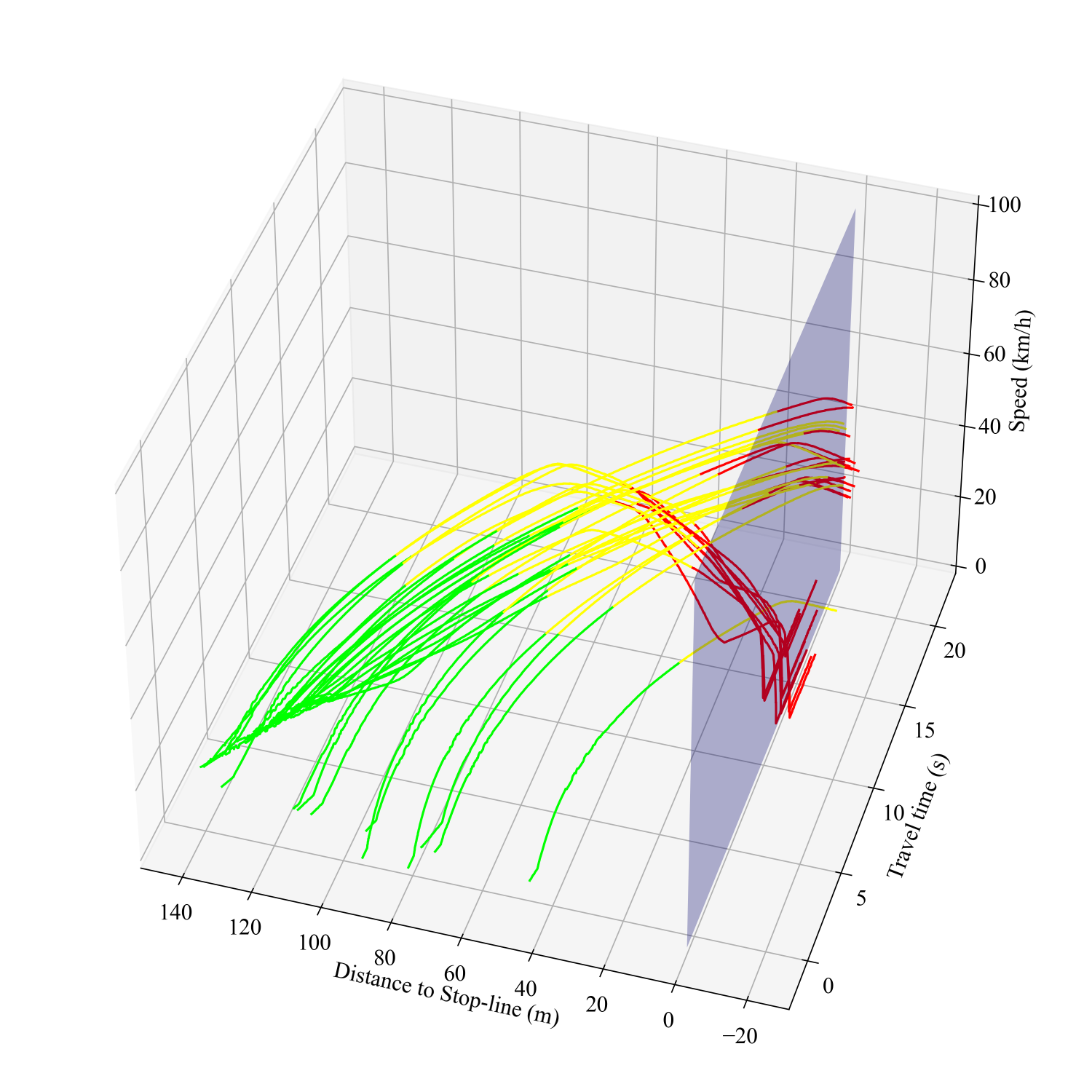}
		\caption{Driver \#1}
		\label{Fig:Driver 1}
\end{subfigure}
\begin{subfigure}{0.49\linewidth}
		\centering
		\includegraphics[width=1.0\linewidth]{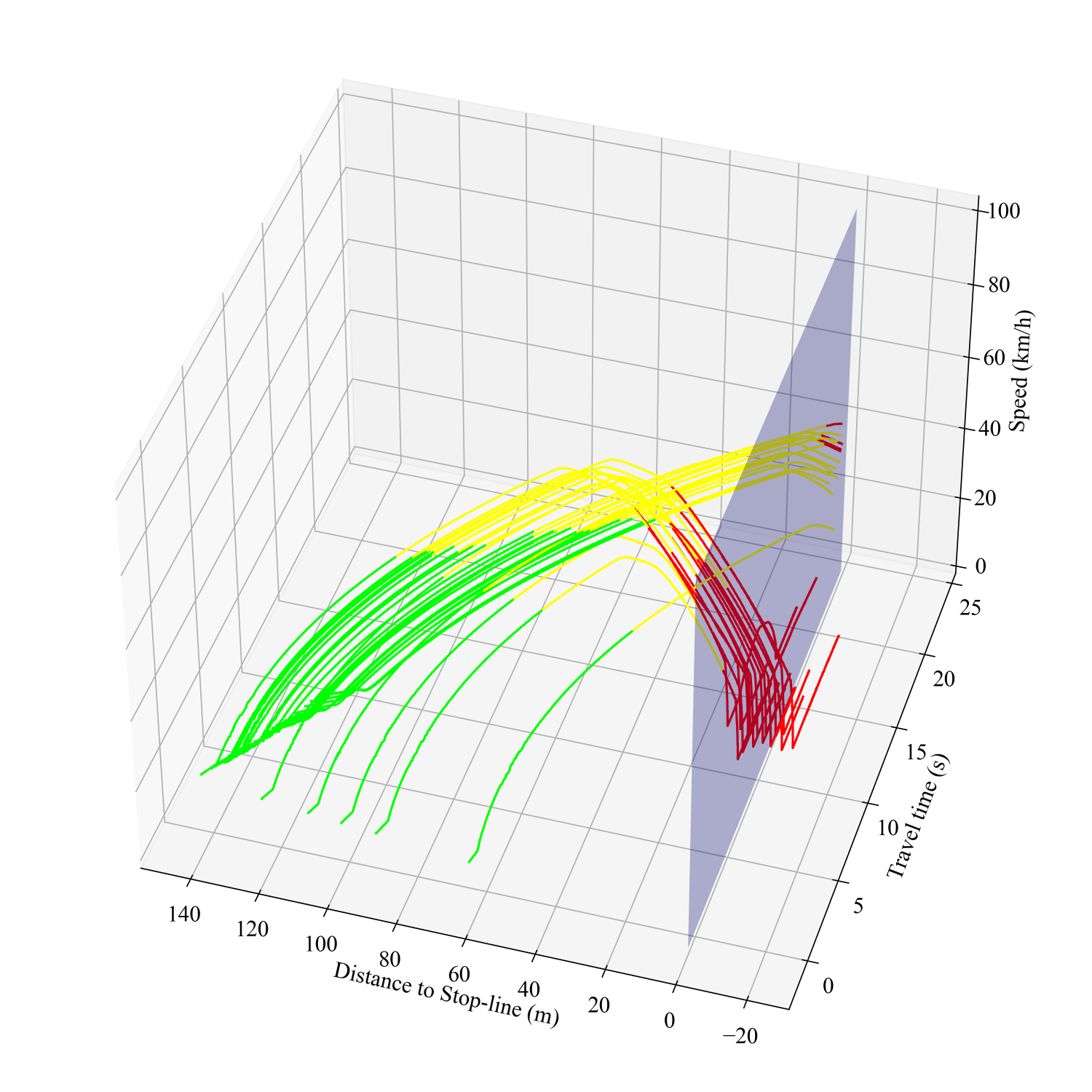}
		\caption{Driver \#2}
		\label{Fig:Driver 2}
\end{subfigure}
\begin{subfigure}{0.49\linewidth}
		\centering
		\includegraphics[width=1.0\linewidth]{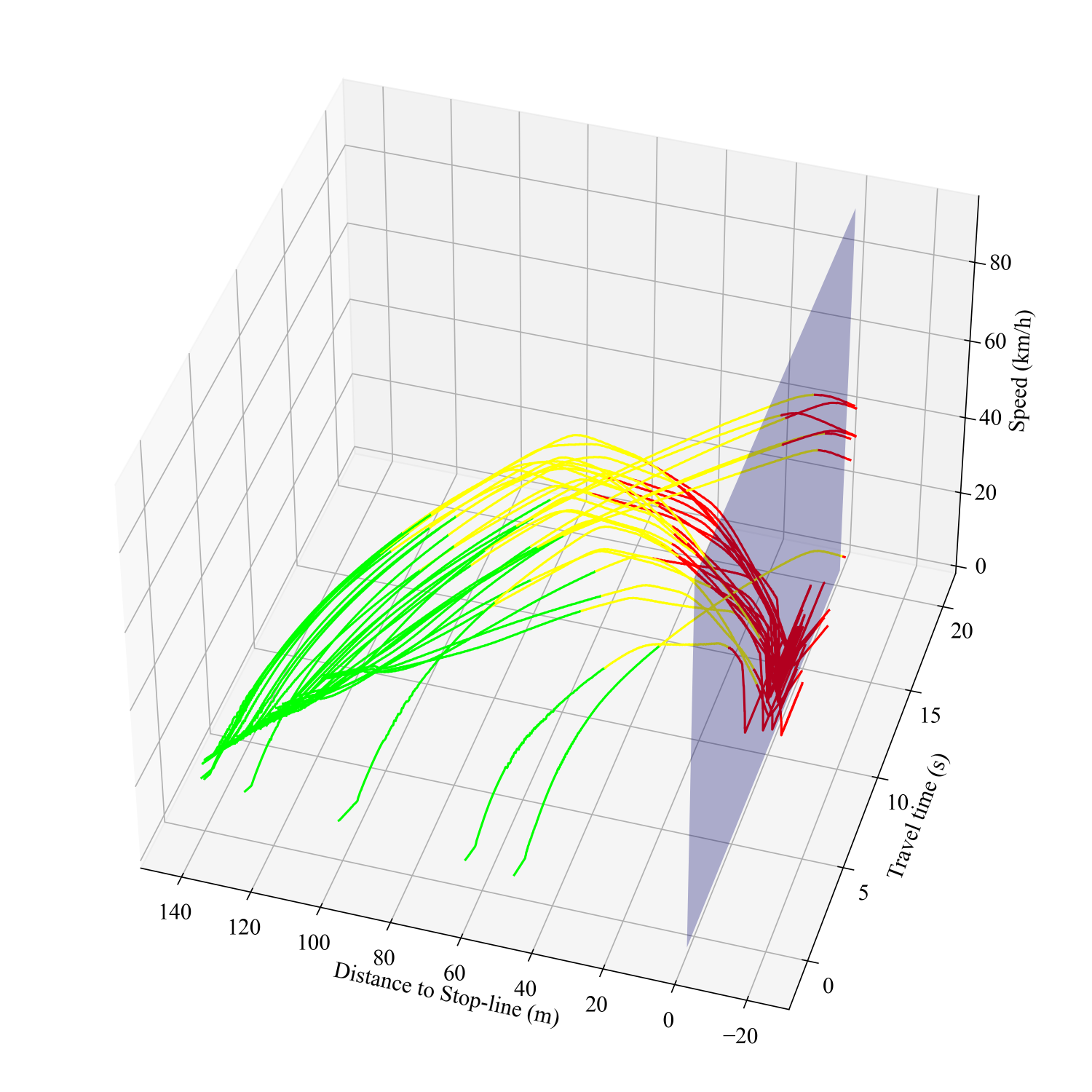}
		\caption{Driver \#3}
		\label{Fig:Driver 3}
\end{subfigure}
\begin{subfigure}{0.49\linewidth}
		\centering
		\includegraphics[width=1.0\linewidth]{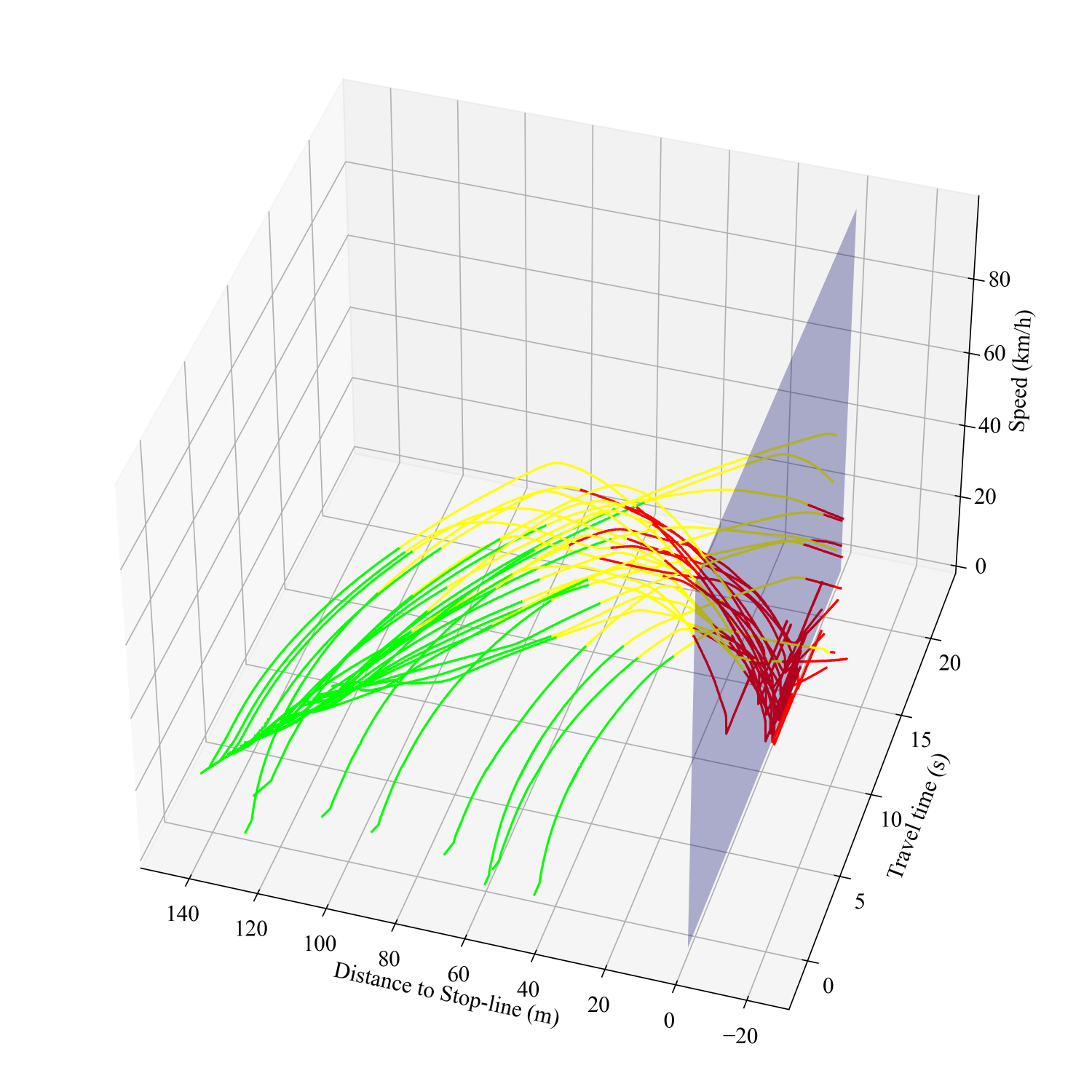}
		\caption{Driver \#4}
		\label{Fig:Driver 4}
\end{subfigure}
\caption{Vehicle trajectories in dilemma zones.}
\label{Fig:Trajectories}
\end{figure}

In analyzing driving behavior in relation to traffic signal changes within the dilemma zone, a clear pattern emerges, as illustrated by the trajectory profiles in Figure \ref{Fig:Trajectories}. The result reveals a common tendency among drivers to accelerate towards their desired speed during the green-light phase, aiming to pass through the intersection before the light changes. Upon the activation of the yellow light, drivers typically do not react instantaneously, suggesting the existence of a reaction time necessary for processing the situation and deciding on the appropriate course of action. Notably, Driver \#1 exhibits a more aggressive driving style, characterized by consistently high speeds near the stop-line and multiple instances of red-light running. This behavior suggests a preference for choosing to pass through the dilemma zone rather than stopping. Driver \#2 and Driver \#4 exhibit more cautious and stable driving behaviors, with fewer occurrences of red-light running, indicating a tendency to adhere more closely to traffic regulations. Driver \#3 is more likely to choose ``stop" when encountering dilemma zones. The observed variability in vehicle trajectories reveals the heterogeneity of driving behavior within the dilemma zone. This diversity illuminates the imperative to integrate personalized characteristics in modeling drivers' decision-making processes.

\subsection{In-depth Investigation of Driver Behavior during the Yellow-light}
\begin{figure}[!ht]
\centering
\begin{subfigure}{\linewidth}
		\centering
		\includegraphics[width=1.0\linewidth]{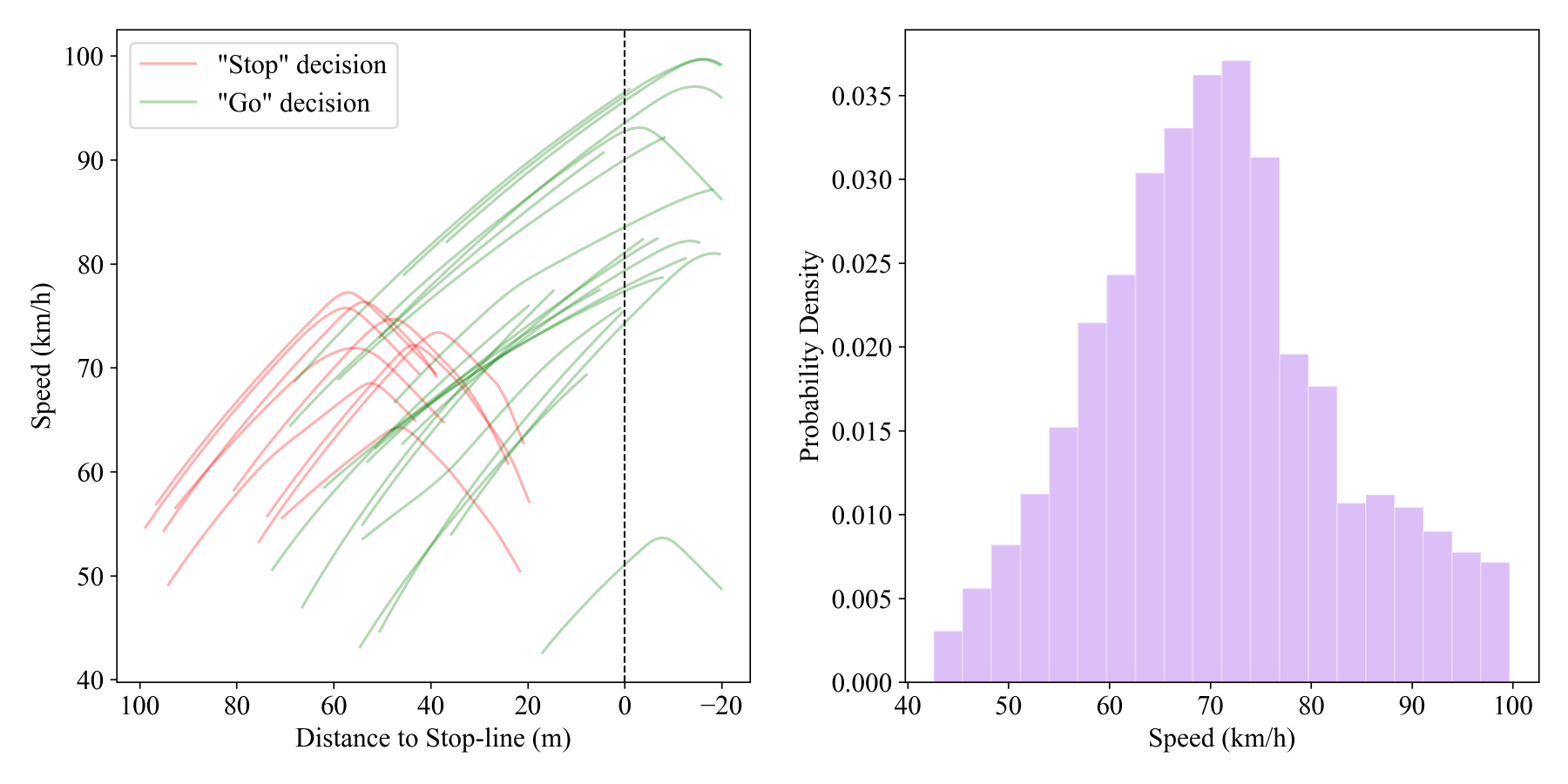}
		\caption{Driver \#1}
		\label{Fig:speed 1}
\end{subfigure}
\begin{subfigure}{\linewidth}
		\centering
		\includegraphics[width=1.0\linewidth]{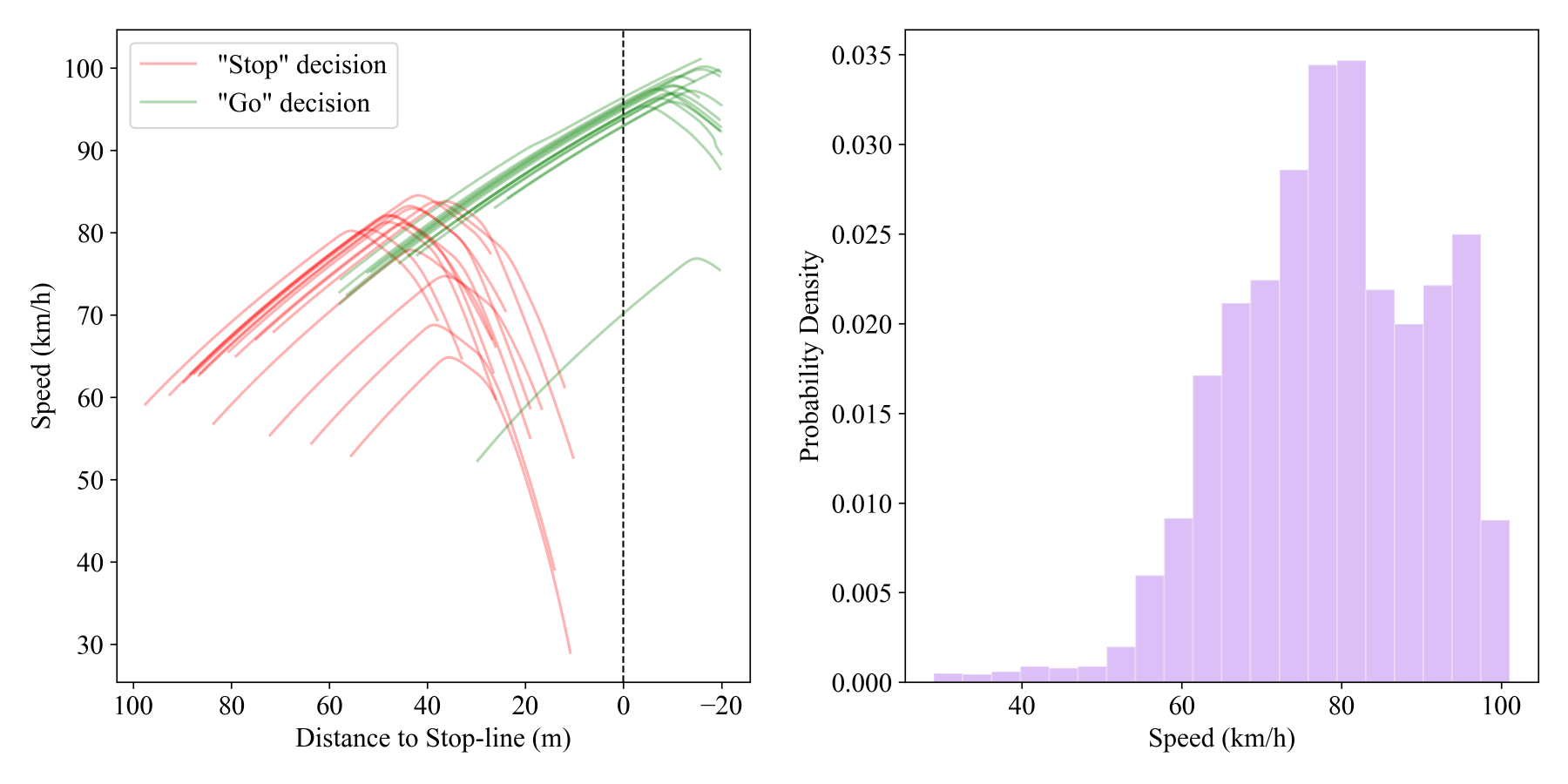}
		\caption{Driver \#2}
		\label{Fig:speed 2}
\end{subfigure}
\begin{subfigure}{\linewidth}
		\centering
		\includegraphics[width=1.0\linewidth]{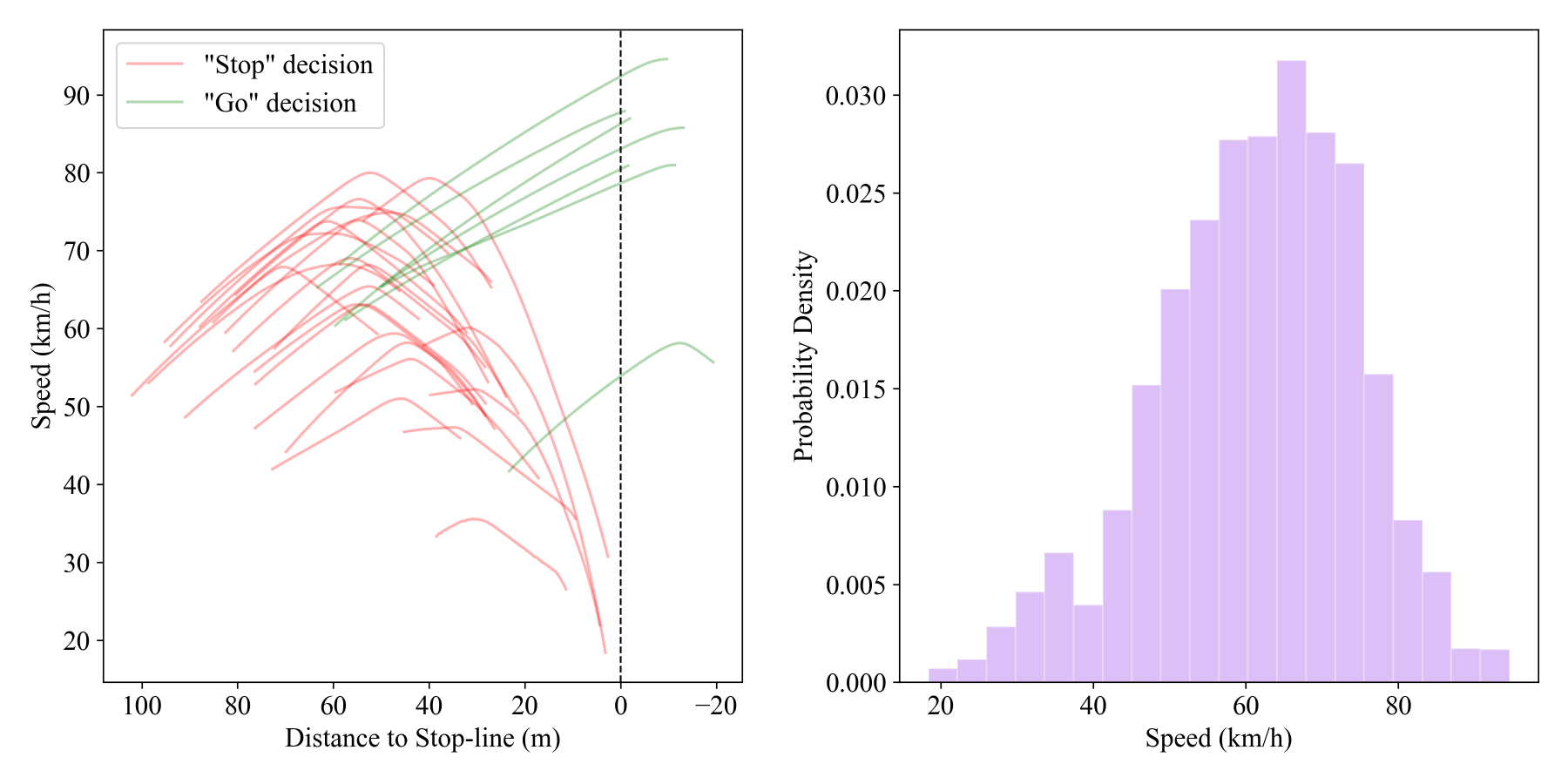}
		\caption{Driver \#3}
		\label{Fig:speed 3}
\end{subfigure}
\begin{subfigure}{\linewidth}
		\centering
		\includegraphics[width=1.0\linewidth]{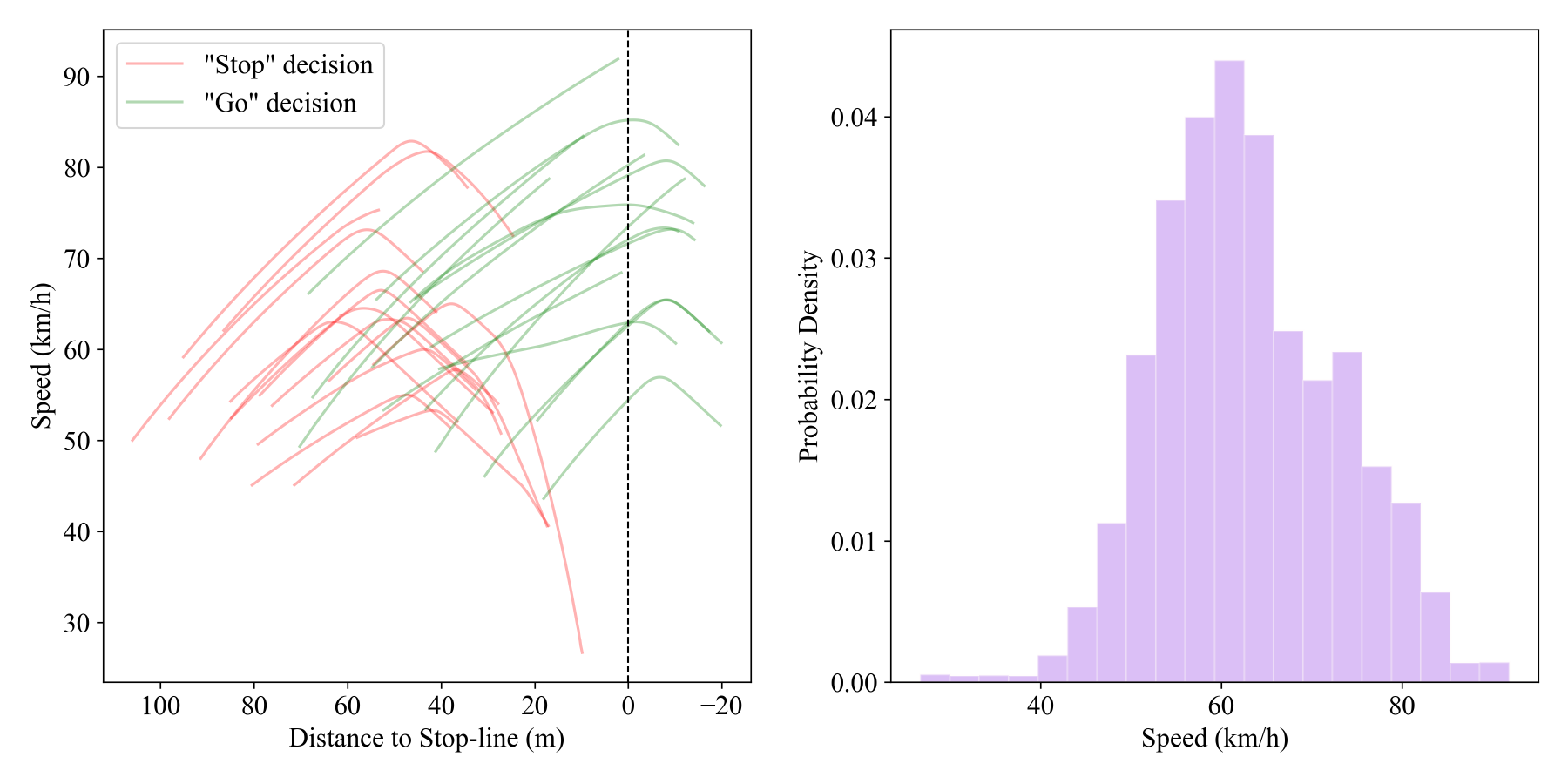}
		\caption{Driver \#4}
		\label{Fig:speed 4}
\end{subfigure}
\caption{Distribution of speed profiles during yellow-light.}
\label{Fig:speed profiles}
\end{figure}

The distribution of vehicle speeds during different phases of the traffic signals, as illustrated in Figure \ref{Fig:speed profiles}, underscores that drivers require a variable amount of actuation time to decelerate upon deciding to stop. The left-hand panel distinctly illustrates that even when a decision to ``stop" is made, drivers require a certain amount of actuation time to decelerate. This actuation time also seems to be variable in different scenarios. Driver \#1 seldom chooses to ``stop" within 70m of the stop-line, in contrast to Driver \#3, who demonstrates a willingness to decelerate even at 40m away. This discrepancy presents the diverse risk assessments and decision-making strategies employed by drivers. Furthermore, the observed consistent behavior among drivers involves accelerating through the stop-line upon selecting the ``go" decision. This strategy appears to be widely adopted with the aim of minimizing the time spent within the intersection and reduce the risk of violations associated with red-light running. The maximum speed also varies between drivers as shown in the right histograms (i.e., approximately 100km/h for Driver \#1 and \#2, compared to around 90km/h for Driver \#3 and \#4). It is also evident that drivers employ varying decision-making strategies, even when their initial states are similar. These inconsistencies imply that the drivers' stop-or-go decision-making process may be influenced by certain latent factors that need to be further investigated in future studies. The histograms on the right provide an aggregate depiction of the distribution of drivers' speeds over 30 trips. These histograms predominantly exhibit unimodal distributions, indicating a tendency towards a single predominant speed among drivers. Nevertheless, it is also evident that there are significant differences in drivers' speed preferences, highlighting the diversity in driving behaviors. 

\subsection{Exploring Drivers' Decision-Making Processes}
Building on the observed heterogeneity in driving behavior and vehicle trajectories, Table \ref{tab: performance} provides a summary of driver behaviors in response to the yellow-light. It includes metrics such as the probability of choosing ``go" decision, the probability of running red-light, as well as average speed, average distance to stop-line, and average yellow time at decision-making moment. It is evident that there are difference in drivers' decision-making within the dilemma zone. Specifically, Driver \#1 is more inclined to proceed ``go" when confronted with a yellow light, concurrently exhibiting an increased likelihood of incurring a red-light violation. 
The 53.3\% probability of Driver \#1 running a red light arises because the stop position often crosses the stop-line at red-light onset, counting as a violation even if the car isn't fully in the intersection. While both Driver \#2 and Driver \#4 have an identical 50.0\% probability of choosing ``go" decision, there exists a substantial divergence in their respective probabilities of running a red-light. It is noteworthy that an aggregate increase in speed does not necessarily correlate with a heightened risk of red-light violations, as illustrated by the cases of Driver \#1 and Driver \#3. The variables presented do not demonstrate a clear correlation with the stop-or-go decision, thereby making the modeling of these variables a critical issue. The Avg. YT, which includes reaction time and decision time, demonstrates minimal variation across the performances of drivers. This consistency can be attributed to the uniform level of focus maintained by all drivers throughout the data collecting phase. 

\begin{table}[ht]
\caption{Driving behaviors in dilemma zone}
\label{tab: performance}
\centering
\begin{tabular}{|c|c|c|c|c|c|}
\hline
   Drivers    & PofGo & PofRR   & Avg. Spd & Avg. DTS & Avg. YT \\\hline
\#1     &   70.0\%                          &   53.3\%                                      &       66.7km/h                                              &   48.0m & 0.8s                                                \\\hline
\#2     &  50.0\%                           &    13.3\%                                     &        77.8km/h                                              &     43.3m &1.0s                                               \\\hline
\#3     &      23.3\%                       &     23.3\%                                    &           62.1km/h                                           &      55.6m &  0.8s                                            \\\hline
\#4     &       50.0\%                      &     30.0\%                                    &                 62.3km/h                                     &          46.6m &1.0s                                        \\\hline
\end{tabular} \\
\vspace{0.2cm}
\footnotesize{*PofGo = Probability of choosing ``go" decision; PofRR = Probability of Running Red-light; Avg. Spd = Average speed at the decision-making moment; Avg. DTS = Average distance to stop-line at the decision-making moment; Avg. YT = Average yellow time at the decision-making moment.}
\end{table}

\subsection{Reconceptualizing Time to Stop-line Metrics}
The time to the stop-line is a critical factor in determining a driver's ability to successfully navigate through a dilemma zone, thereby serving as a criterion to delineate the boundaries of Type II dilemma zone \cite{bonneson2002intelligent}. However, the conventional definition of time to stop-line appears to be oversimplified, as it merely calculates the time required for a driver to reach the stop-line at a constant speed upon encountering the dilemma zone. In fact, drivers gauge their ability to stop at the stop-line by estimating their maximum deceleration potential when choosing the ``stop" decision, and similarly, they assess their maximum acceleration ability to pass through the intersection when selecting the ``go" decision. To more realistically portray the effect of time to stop-line on driver decision-making, this study incorporates vehicle dynamics (i.e., constant acceleration/deceleration model) for a refined estimation, as shown in \autoref{eq1} and \autoref{eq2}.

\textit{Time to stop-line (given the decision of ``stop" is made)}
\begin{equation}
    \label{eq1}
    v_0 = bt_b
\end{equation}

\textit{Time to stop-line (given the decision of ``go" is made)}
\begin{equation}
    \label{eq2}
    x = v_0t_a+\frac{1}{2}at_a^2
\end{equation}

where $x$ is the distance to stop-line, $v_0$ is the current speed, $a=3m/s^2$ and $b=-3m/s^2$ are the maximum acceleration and maximum deceleration rates, respectively. The estimated travel time (i.e., time to stop-line: $t_b$ and $t_a$) can be obtained when the driver makes decisions.

\begin{figure}[!ht]
\centering
\begin{subfigure}{\linewidth}
		\centering
		\includegraphics[width=1.0\linewidth]{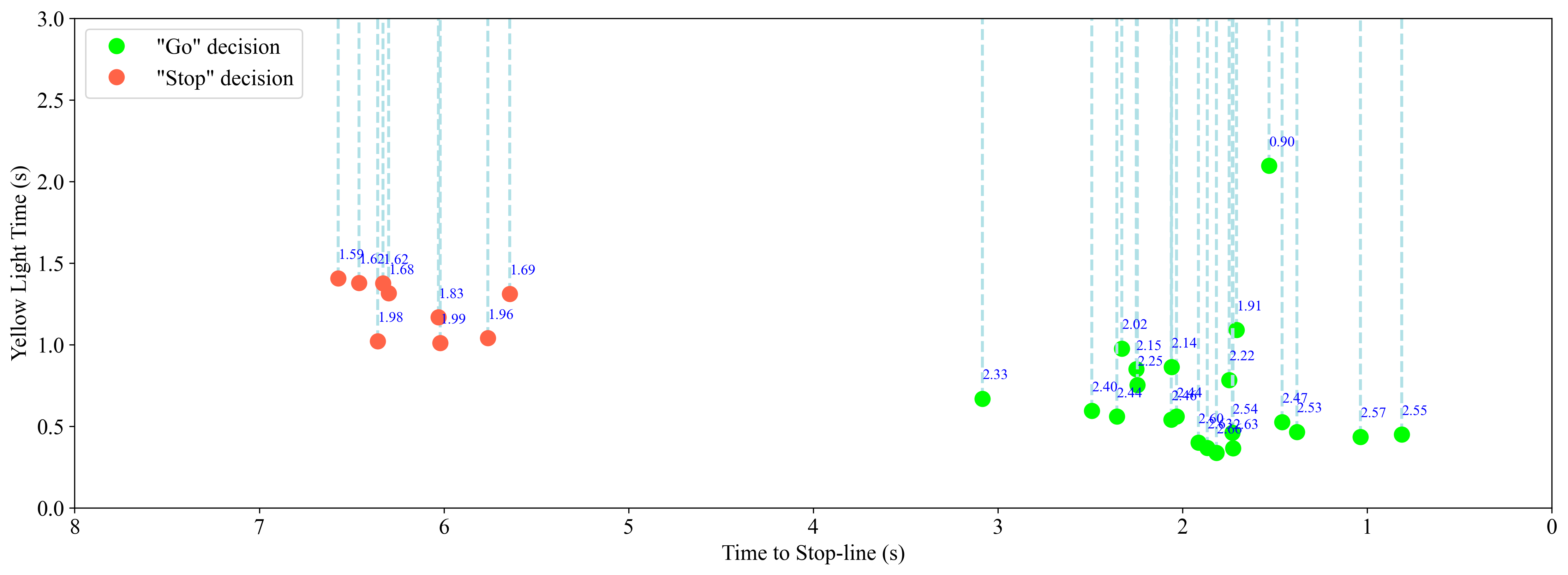}
		\caption{Driver \#1}
		\label{Fig:time 1}
\end{subfigure}
\begin{subfigure}{\linewidth}
		\centering
		\includegraphics[width=1.0\linewidth]{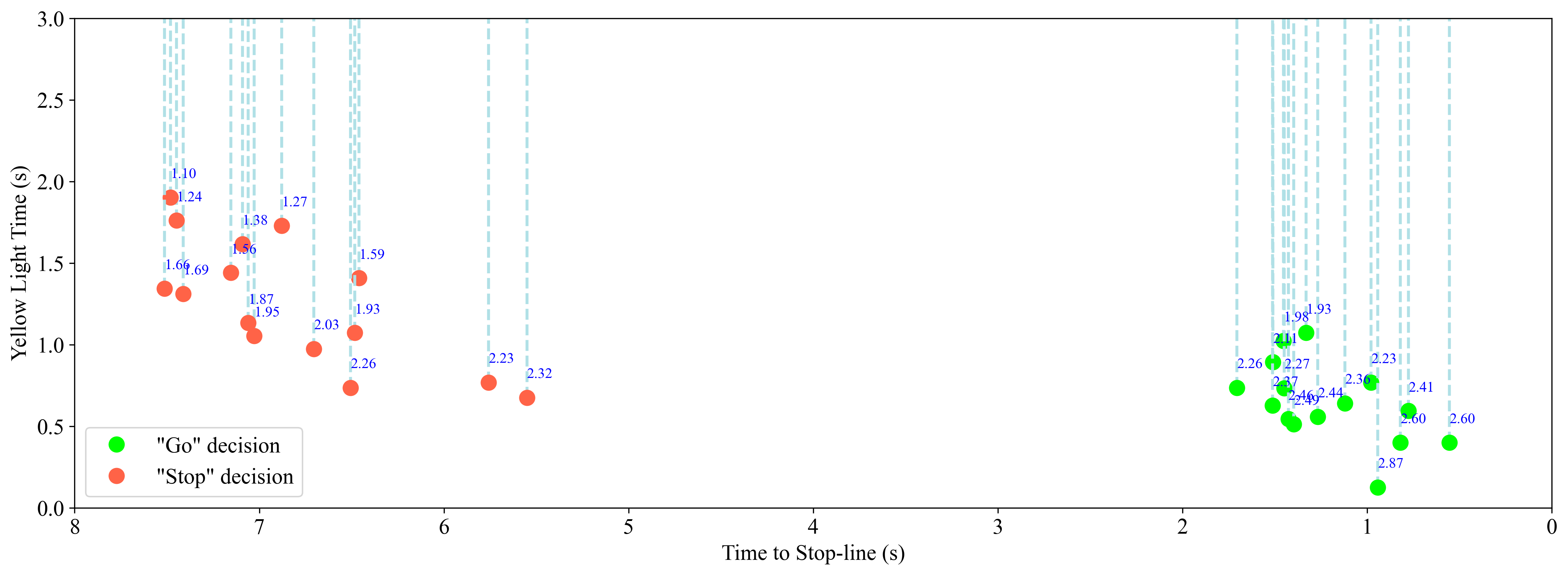}
		\caption{Driver \#2}
		\label{Fig:time 2}
\end{subfigure}
\begin{subfigure}{\linewidth}
		\centering
		\includegraphics[width=1.0\linewidth]{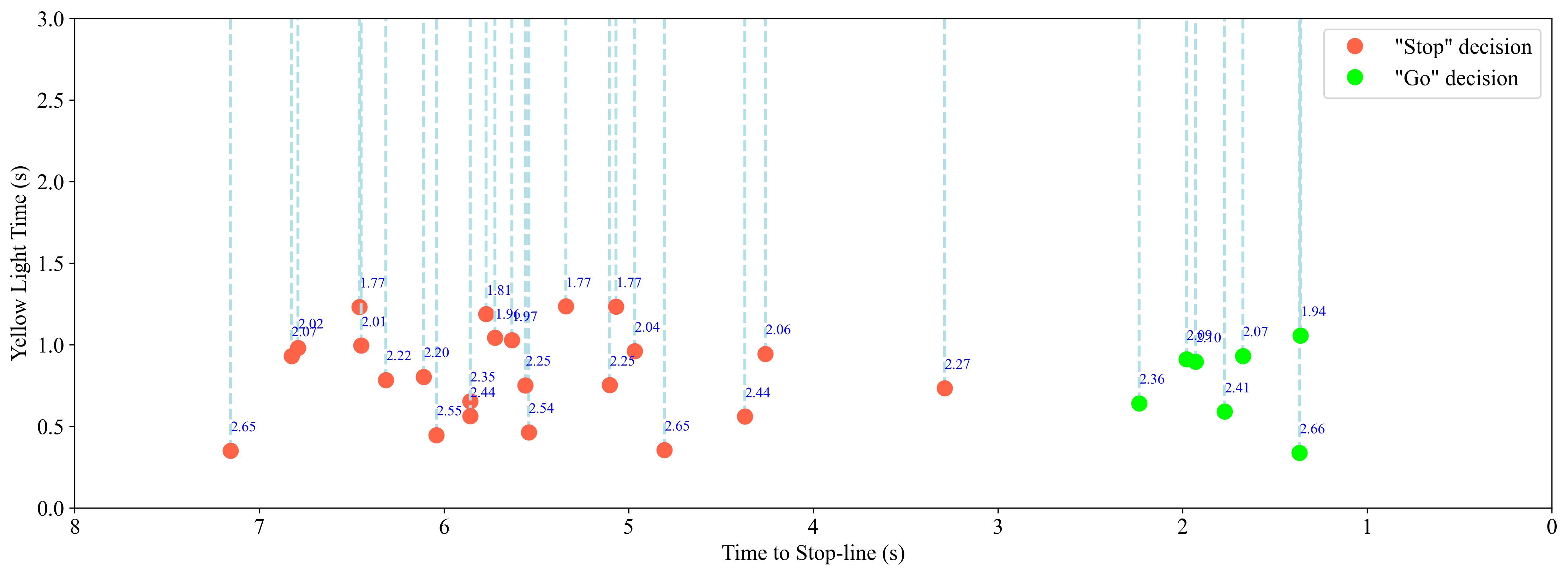}
		\caption{Driver \#3}
		\label{Fig:time 3}
\end{subfigure}
\begin{subfigure}{\linewidth}
		\centering
		\includegraphics[width=1.0\linewidth]{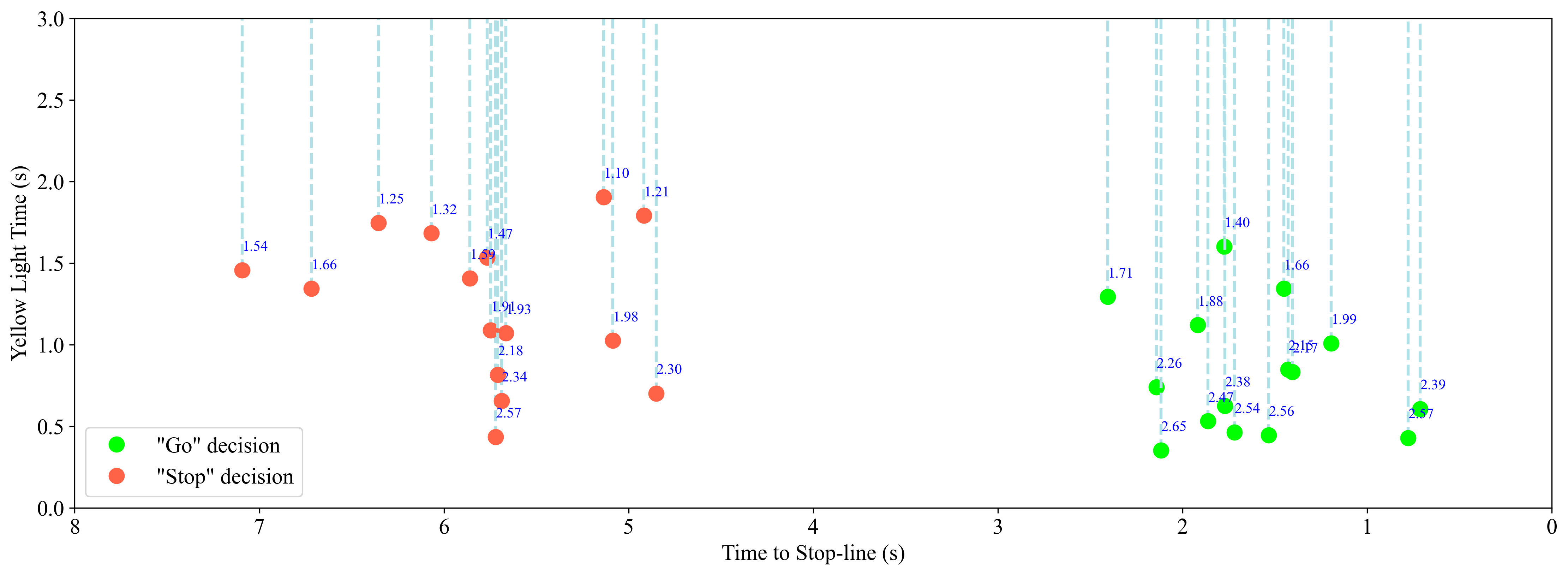}
		\caption{Driver \#4}
		\label{Fig:time 4}
\end{subfigure}
\caption{Moments of ``stop-or-go" decision-making by drivers.}
\label{Fig:time}
\end{figure}

Utilizing the time to stop-line estimation methodology previously outlined, we calculated the moments at which drivers make the decision to ``stop" or ``go" as shown in Figure \ref{Fig:time}. It is evident that the estimated time to reach the stop-line is longer for the ``stop" decision scenarios compared to the ``go" decision scenarios. Furthermore, variations are observable in the timing of decisions across different drivers. Specifically, the ``stop" decision intervals for Driver \#1 and Driver \#4 are between 5 and 7 seconds to the stop-line. Driver \#2's decisions mainly fall within a 6 to 8 second range, while Driver \#3 demonstrates a boarder range of decision timings. In scenarios where ``go" is chosen, Driver \#2's decision timing falls within 2 seconds of the estimated time to reach the stop-line, while the other drivers' decisions are generally made within a 2.5-second window. It is also found that, from the onset of the yellow light, the decision to choose ``go" generally occurs earlier than the decision to ``stop". In addition, as indicated in the figure, when selecting ``stop", the remaining time of yellow light is longer than the time needed to reach the stop-line, implying that the driver must decelerate to arrive at the stop-line before the light turns red. When choosing to ``go", if the remaining time of the yellow light exceeds the time to reach the stop-line, it indicates that the driver can proceed without running the red light. 

\section{Personalized stop-or-go prediction model}
\label{sec: prediction}
The analysis conducted in previous sections has uncovered significant insights into driving behavior within the dilemma zone, demonstrating not only variability between different drivers (inter-driver heterogeneity) but also variations in decision-making by the same driver across similar traffic scenarios (intra-driver heterogeneity) \cite{10421850}. In response to these findings, this section introduces a personalized stop-or-go prediction model. This model is specifically tailored to account for the unique driving patterns of individual drivers, aiming to accurately describe and predict their behavior within the dilemma zone. 

\subsection{Structure of the Personalized Transformer Encoder}
The conventional approach for predicting stop-or-go decisions in dilemma zones primarily considers the state at the onset of the yellow-light or vehicle trajectories during the yellow-light interval, utilizing a binary logistic regression model \cite{chauhan2022analysing}. Significantly, the configuration of the utility function have a substantial influence on the prediction results. The ability of neural networks to process time series data and multiple influential factors holds potential for enhancing decision in dilemma zones. Compared to the LSTM, the self-attention mechanism of Transformer can process all sequence information in parallel \cite{Shi_2023_ICCV}. Moreover, it has the capability to focus on crucial information within the sequence, making it suitable for predicting stop-or-go decisions in dilemma zones. This is because such decisions always depend on the state at specific moments within the entire sequence. By accurately capturing the driver's characteristics, this study introduces a Personalized Transformer Encoder to predict the driver's decision-making process in the dilemma zone. The model assimilates inputs comprising both common and personalized information. The common information includes vehicle speed, distance to stop-line, and current yellow light time. Meanwhile, the personalized information encompasses statistical information specific to each driver, such as the average speed at the decision-making moment, average distance to stop-line at the decision-making moment, and average yellow time at the decision-making moment, as detailed in Table \ref{tab: performance}.

\begin{figure}[!ht]
\centering
\includegraphics[width=0.92\linewidth]{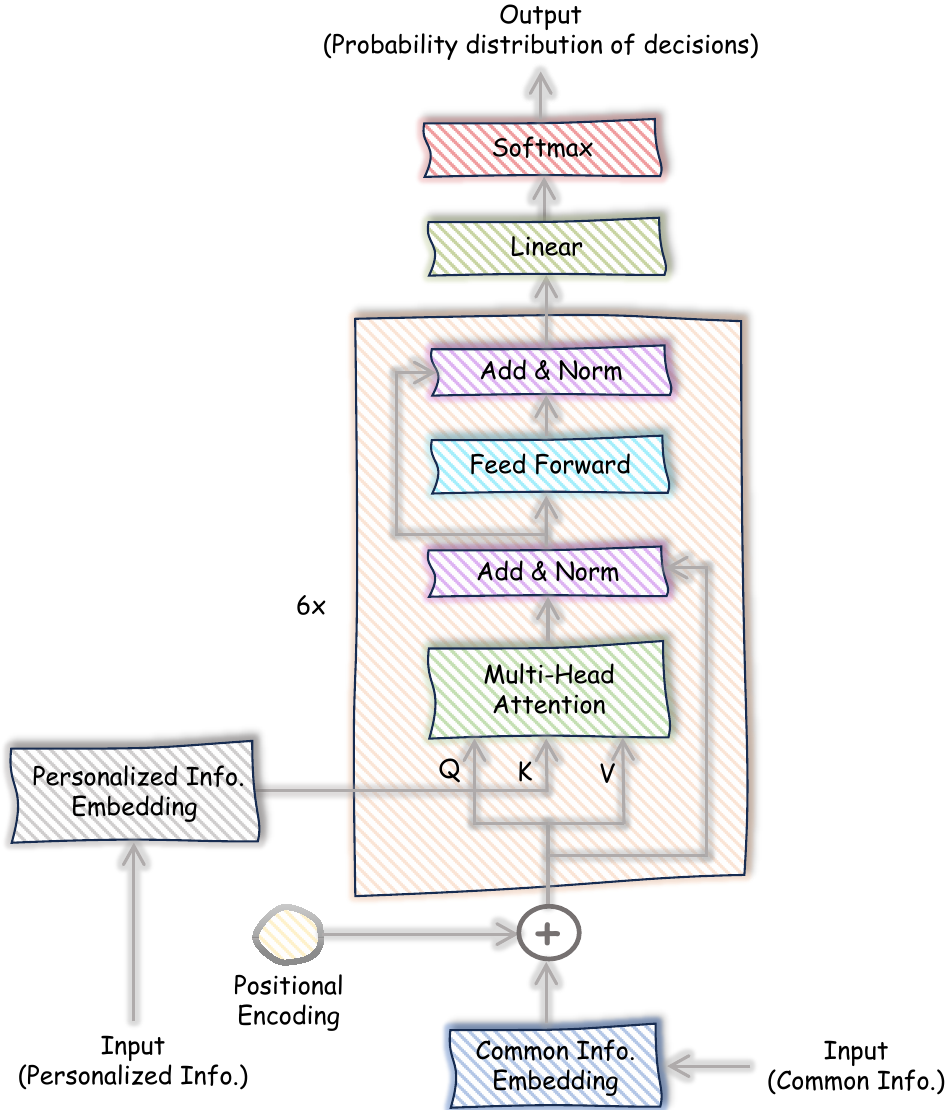}
\caption{Personalized Transformer Encoder for decision-making prediction.}
\label{Fig:nn}
\end{figure}

As shown in Figure \ref{Fig:nn}, the common information is utilized to compute the query (Q) and value (V) matrices, while personalized information is used to compute the key (K) matrix in the multi-head attention layer. This design is inspired by the role of personalized information, which reflects the drivers' state at the decision-making moment. Q and K are are multiplied to determine the similarity scores. A higher similarity between Q and K suggests that the information is more relevant to the drivers' decision-making process, thereby resulting in an higher attention score.

\subsection{Results}

To demonstrate the advanced capabilities of the proposed Personalized Transformer Encoder, this study conducts a comparative analysis against the binary logistic regression model and the Generic Transformer Encoder. The two Transformer Encoders share the same network structure, and the key difference is that the Generic Transformer Encoder does not incorporate personalized information to calculate K, relying solely on common information during the training process. The findings, as presented in Table \ref{tab:improve}, reveal that the personalized prediction model exhibits enhanced prediction accuracy over the baseline model. This enhancement is observed across both individual driver assessments and within a randomly segmented test set. Specifically, the improvement in prediction accuracy ranging from 3.7\% to 12.6\% when compared to the Generic Transformer Encoder, and from 16.8\% to 21.6\% over the binary logistic regression model. This comparison highlights the significant advantage of incorporating personalized data in improving the predictive capability of driver decision-making models.

\begin{table}[ht]
\caption{Comparison of prediction accuracy}
\label{tab:improve}
\begin{tabular}
{|c|c|c|c|c|c|}
\hline
                   & Driver \#1 & Driver \#2 & Driver \#3 & Driver \#4 & R. \\ \hline
{\color[HTML]{00D2CB}B.L.R.} & 76.7\%     & 73.3\%     & 80.0\%     & 76.7\%     & 80.3\% \\ \hline                   
{\color[HTML]{FD6864}G.T.} & 90.0\%     & 80.0\%     & 90.0\%     & 86.7\%     & 83.3\% \\ \hline
P.T.     & 93.3\%     & 86.7\%     & 96.7\%     & 93.3\%     & 93.8\% \\ \hline
\multicolumn{1}{|c|}{\multirow{2}{*}{IMPRV.}} & \multicolumn{1}{c|}{{\color[HTML]{00D2CB}21.6\%$\uparrow$}} & \multicolumn{1}{c|}{{\color[HTML]{00D2CB}18.3\%$\uparrow$}} & \multicolumn{1}{c|}{{\color[HTML]{00D2CB}20.9\%$\uparrow$}} & \multicolumn{1}{c|}{{\color[HTML]{00D2CB}21.6\%$\uparrow$}} & \multicolumn{1}{c|}{{\color[HTML]{00D2CB}16.8\%$\uparrow$}} \\ \cline{2-6} 
\multicolumn{1}{|c|}{}                       & \multicolumn{1}{c|}{{\color[HTML]{FD6864}3.7\%$\uparrow$}} & \multicolumn{1}{c|}{{\color[HTML]{FD6864}8.4\%$\uparrow$}} & \multicolumn{1}{c|}{{\color[HTML]{FD6864}7.4\%$\uparrow$}} & \multicolumn{1}{c|}{{\color[HTML]{FD6864}7.6\%$\uparrow$}} & \multicolumn{1}{c|}{{\color[HTML]{FD6864}12.6\%$\uparrow$}} \\ \hline
\end{tabular}\\

\footnotesize{*B.L.R. = Binary Logistic Regression Model; G.T. = Generic Transformer; P.T. = Personalized Transformer; IMPRV. = Improvement; R. = Random.}
\end{table}

\section{Conclusions and future work}
\label{sec: con}
Dilemma zones at intersections pose considerable safety and efficiency challenges, marked by drivers' inconsistent and sometimes inappropriate decisions. The study of drivers' decision-making within dilemma zones is compelling due to the diverse behaviors observed, with variability both across different drivers and within an individual driver's repeated encounters with the same scenario. This complexity highlights the challenge in predicting driver behavior in dilemma zones, where the decision to ``stop" or ``go" can be influenced by various factors. Understanding and modeling these diverse decision-making processes are crucial for enhancing traffic safety, efficiency, and environmental sustainability. To address this, we developed a CARLA-enabled driving simulator to gather real-time, high-resolution data on traffic signal information, vehicle trajectories, and drivers' stop-or-go decisions. Through statistical analysis, we captured personalized information, which served as the foundation for developing a tailored model to predict driver decision. This model, integrating both the personalized information and the common information, has been experimentally shown to enhance prediction accuracy by 3.7\% to 12.6\% compared to the Generic Transformer Encoder, and from 16.8\% to 21.6\% over the binary logistic regression model. The proposed personalized decision prediction model offers significant benefits to autonomous vehicles and traffic operators by enabling the real-time prediction of other vehicles' decisions based on their characteristics within the dilemma zone. Such insights are crucial for developing personalized advanced driver assistance systems (ADAS), enabling adaptive adjustments to vehicle speeds and traffic signal timings, thereby improving safety and efficiency.

Future endeavors will focus on expanding the dataset to encompass more dilemma zone scenarios, including variations in yellow-light duration and traffic flow conditions, alongside incorporating real-world roadside sensors data. Concurrently, we plan to recruit more volunteer drivers to facilitate a more comprehensive investigation into personalized driving behavior in dilemma zones. 

\bibliographystyle{IEEEtran} 
\bibliography{IEEEabrv,IEEEexample}

\begin{thebibliography}{10}
\providecommand{\url}[1]{#1}
\csname url@rmstyle\endcsname
\providecommand{\newblock}{\relax}
\providecommand{\bibinfo}[2]{#2}
\providecommand\BIBentrySTDinterwordspacing{\spaceskip=0pt\relax}
\providecommand\BIBentryALTinterwordstretchfactor{4}
\providecommand\BIBentryALTinterwordspacing{\spaceskip=\fontdimen2\font plus
\BIBentryALTinterwordstretchfactor\fontdimen3\font minus \fontdimen4\font\relax}
\providecommand\BIBforeignlanguage[2]{{%
\expandafter\ifx\csname l@#1\endcsname\relax
\typeout{** WARNING: IEEEtran.bst: No hyphenation pattern has been}%
\typeout{** loaded for the language `#1'. Using the pattern for}%
\typeout{** the default language instead.}%
\else
\language=\csname l@#1\endcsname
\fi
#2}}

\bibitem{gazis1960problem}
D.~Gazis, R.~Herman, and A.~Maradudin, ``The problem of the amber signal light in traffic flow,'' \emph{Operations Research}, vol.~8, no.~1, pp. 112--132, 1960.

\bibitem{wei2024dilemma}
C.~Wei, Z.~Qin, G.~Wu, M.~J. Barth, A.~Abdelraouf, R.~Gupta, and K.~Han, ``Dilemma zone: A comprehensive study of influential factors and behavior analysis,'' in \emph{2024 Forum for Innovative Sustainable Transportation Systems (FISTS)}.\hskip 1em plus 0.5em minus 0.4em\relax IEEE, 2024, pp. 1--8.

\bibitem{chauhan2022analysing}
R.~Chauhan, A.~Dhamaniya, and S.~Arkatkar, ``Analysing driver’s decision in dilemma zone at signalized intersections under disordered traffic conditions,'' \emph{Transportation Research Part F: Traffic Psychology and Behaviour}, vol.~89, pp. 222--235, 2022.

\bibitem{qin2024game}
Z.~Qin, A.~Ji, Z.~Sun, G.~Wu, P.~Hao, and X.~Liao, ``Game theoretic application to intersection management: A literature review,'' \emph{IEEE Transactions on Intelligent Vehicles}, 2024.

\bibitem{das2022traffic}
D.~Das, N.~V. Altekar, K.~L. Head, and F.~Saleem, ``Traffic signal priority control strategy for connected emergency vehicles with dilemma zone protection for freight vehicles,'' \emph{Transportation Research Record}, vol. 2676, no.~1, pp. 499--517, 2022.

\bibitem{gao2021coordinated}
Y.~Gao, H.~Hu, and Y.~Liu, ``Coordinated arterial dilemma zone protection through dynamic signal timing optimization,'' \emph{IEEE Transactions on Intelligent Transportation Systems}, vol.~23, no.~6, pp. 5434--5445, 2021.

\bibitem{zegeer1978green}
C.~V. Zegeer and R.~C. Deen, \emph{Green-extension systems at high-speed intersections}.\hskip 1em plus 0.5em minus 0.4em\relax Citeseer, 1978, vol. 496.

\bibitem{sayed1999advance}
T.~Sayed, H.~Vahidi, and F.~Rodriguez, ``Advance warning flashers: Do they improve safety?'' \emph{Transportation Research Record}, vol. 1692, no.~1, pp. 30--38, 1999.

\bibitem{zimmerman2007additional}
K.~Zimmerman, ``Additional dilemma zone protection for trucks at high-speed signalized intersections,'' \emph{Transportation Research Record}, vol. 2009, no.~1, pp. 82--88, 2007.

\bibitem{elmitiny2010classification}
N.~Elmitiny, X.~Yan, E.~Radwan, C.~Russo, and D.~Nashar, ``Classification analysis of driver's stop/go decision and red-light running violation,'' \emph{Accident Analysis \& Prevention}, vol.~42, no.~1, pp. 101--111, 2010.

\bibitem{rakha2008modeling}
H.~Rakha, A.~Amer, and I.~El-Shawarby, ``Modeling driver behavior within a signalized intersection approach decision--dilemma zone,'' \emph{Transportation Research Record}, vol. 2069, no.~1, pp. 16--25, 2008.

\bibitem{ghanipoor2016modeling}
S.~Ghanipoor~Machiani and M.~Abbas, ``Modeling human learning and cognition structure: Application to driver behavior in dilemma zone,'' \emph{Journal of Transportation Engineering}, vol. 142, no.~11, p. 04016057, 2016.

\bibitem{papaioannou2007driver}
P.~Papaioannou, ``Driver behaviour, dilemma zone and safety effects at urban signalised intersections in greece,'' \emph{Accident Analysis \& Prevention}, vol.~39, no.~1, pp. 147--158, 2007.

\bibitem{liao2023driver}
X.~Liao, X.~Zhao, Z.~Wang, Z.~Zhao, K.~Han, R.~Gupta, M.~J. Barth, and G.~Wu, ``Driver digital twin for online prediction of personalized lane change behavior,'' \emph{IEEE Internet of Things Journal}, 2023.

\bibitem{zhao2022personalized}
Z.~Zhao, Z.~Wang, K.~Han, R.~Gupta, P.~Tiwari, G.~Wu, and M.~J. Barth, ``Personalized car following for autonomous driving with inverse reinforcement learning,'' in \emph{2022 International Conference on Robotics and Automation (ICRA)}.\hskip 1em plus 0.5em minus 0.4em\relax IEEE, 2022, pp. 2891--2897.

\bibitem{bao2021prediction}
N.~Bao, A.~Carballo, and T.~Kazuya, ``Prediction of personalized driving behaviors via driver-adaptive deep generative models,'' in \emph{2021 IEEE Intelligent Vehicles Symposium (IV)}.\hskip 1em plus 0.5em minus 0.4em\relax IEEE, 2021, pp. 616--621.

\bibitem{Li2024personalized}
S.~Li, C.~Wei, G.~Wu, M.~J. Barth, A.~Abdelraouf, R.~Gupta, and K.~Han, ``Personalized trajectory prediction for driving behavior modeling in ramp-merging scenarios,'' in \emph{7th IEEE International Conference on Robotic Computing}.\hskip 1em plus 0.5em minus 0.4em\relax IEEE, 2023.

\bibitem{vallon2017machine}
C.~Vallon, Z.~Ercan, A.~Carvalho, and F.~Borrelli, ``A machine learning approach for personalized autonomous lane change initiation and control,'' in \emph{2017 IEEE Intelligent vehicles symposium (IV)}.\hskip 1em plus 0.5em minus 0.4em\relax IEEE, 2017, pp. 1590--1595.

\bibitem{huang2021personalized}
C.~Huang, H.~Huang, P.~Hang, H.~Gao, J.~Wu, Z.~Huang, and C.~Lv, ``Personalized trajectory planning and control of lane-change maneuvers for autonomous driving,'' \emph{IEEE Transactions on Vehicular Technology}, vol.~70, no.~6, pp. 5511--5523, 2021.

\bibitem{butakov2016personalized}
V.~A. Butakov and P.~Ioannou, ``Personalized driver assistance for signalized intersections using v2i communication,'' \emph{IEEE Transactions on Intelligent Transportation Systems}, vol.~17, no.~7, pp. 1910--1919, 2016.

\bibitem{abdelraouf2023interactionaware}
A.~Abdelraouf, R.~Gupta, and K.~Han, ``Interaction-aware personalized vehicle trajectory prediction using temporal graph neural networks,'' in \emph{2023 IEEE 26th International Conference on Intelligent Transportation Systems (ITSC)}.\hskip 1em plus 0.5em minus 0.4em\relax IEEE, 2023, pp. 2070--2077.

\bibitem{dosovitskiy2017carla}
A.~Dosovitskiy, G.~Ros, F.~Codevilla, A.~Lopez, and V.~Koltun, ``Carla: An open urban driving simulator,'' in \emph{Conference on robot learning}.\hskip 1em plus 0.5em minus 0.4em\relax PMLR, 2017, pp. 1--16.

\bibitem{bonneson2002intelligent}
J.~Bonneson, D.~Middleton, K.~Zimmerman, H.~Charara, and M.~Abbas, ``Intelligent detection-control system for rural signalized intersections,'' \emph{Texas Department of Transportation}, 2002.

\bibitem{savolainen2016driver}
P.~T. Savolainen, A.~Sharma, and T.~J. Gates, ``Driver decision-making in the dilemma zone--examining the influences of clearance intervals, enforcement cameras and the provision of advance warning through a panel data random parameters probit model,'' \emph{Accident Analysis \& Prevention}, vol.~96, pp. 351--360, 2016.

\bibitem{10421850}
X.~Yao, S.~C. Calvert, and S.~P. Hoogendoorn, ``Identification of driving heterogeneity using action-chains,'' in \emph{2023 IEEE 26th International Conference on Intelligent Transportation Systems (ITSC)}, 2023, pp. 6001--6006.

\bibitem{Shi_2023_ICCV}
L.~Shi, L.~Wang, S.~Zhou, and G.~Hua, ``Trajectory unified transformer for pedestrian trajectory prediction,'' in \emph{Proceedings of the IEEE/CVF International Conference on Computer Vision (ICCV)}, October 2023, pp. 9675--9684.

\end{thebibliography}

\end{document}